\documentclass{article}

\PassOptionsToPackage{numbers, compress}{natbib}

 \usepackage[preprint]{neurips_2024}

\usepackage[utf8]{inputenc} 
\usepackage[T1]{fontenc}    
\usepackage{hyperref}       
\usepackage{url}            
\usepackage{booktabs}       
\usepackage{amsfonts}       
\usepackage{nicefrac}       
\usepackage{microtype}      
\usepackage{xcolor}         
\usepackage{colortbl}
\usepackage{graphicx}
\usepackage{amsmath}
\usepackage{amssymb}
\usepackage{dsfont}

\definecolor{LightCyan}{rgb}{0.75,0.95,1}

\definecolor{gr}{rgb}{0,0.784,0.313}
\definecolor{lightb}{rgb}{0.05,0.48,1}
\definecolor{darkb}{rgb}{0.007,0,1}
\definecolor{purp}{rgb}{0.462,0,0.988}

\newcolumntype{a}{>{\columncolor{LightCyan}}c}

\title{OW-VISCapTor: Abstractors for Open-World Video Instance Segmentation and Captioning}

\author{
Anwesa Choudhuri, Girish Chowdhary and Alexander G. Schwing \\
 University of Illinois at Urbana-Champaign \\
\texttt{\{anwesac2,girishc,aschwing\}@illinois.edu} \\
}

\begin{document}

\maketitle

\begin{abstract}

We propose the new task `open-world video instance segmentation \emph{and} captioning'. It requires to detect, segment, track \emph{and} describe with rich captions never before seen objects. This challenging task can be addressed by developing ``abstractors'' which connect a vision model and a language foundation model. Concretely, we connect a multi-scale visual feature extractor and a large language model (LLM) by developing an object abstractor and an object-to-text abstractor. 
The object abstractor, consisting of a prompt encoder and transformer blocks, introduces spatially-diverse open-world object queries to discover never before seen objects in videos. An inter-query contrastive loss further encourages the diversity of object queries. The object-to-text abstractor is augmented with masked cross-attention and acts as a bridge between the object queries and a frozen LLM to generate rich and descriptive object-centric captions for each detected object. Our generalized approach surpasses the baseline that jointly addresses the tasks of open-world video instance segmentation and dense video object captioning by $13\%$ on never before seen objects, and by $10\%$ on object-centric captions. 

\end{abstract}

\section{Introduction}
\label{sec:intro}

We propose the generalized task of open-world video instance segmentation \emph{and} captioning (OW-VISCap). This task combines open-world video instance segmentation (OW-VIS)~\cite{owvisformer, tao, burst, uvo, owtb} and the generation of rich object-centric captions for objects~\cite{densevoc2023}. Specifically, OW-VISCap involves detecting, segmenting, and tracking previously seen or unseen objects in a video, while simultaneously generating free-form captions for each of the detected/segmented objects. 
The open-world aspect makes this task widely applicable. 
However, it also makes 
this task  challenging because the objects are often never seen during training, are occasionally partly or entirely occluded, the appearance and position of these objects change over time, the objects may leave the scene only to re-appear at a later time, and because generating free-form object-centric captions requires a powerful object representation as well as a strong natural language understanding. 
Addressing these challenges to obtain an accurate method for this task that works online is crucial in fields like  autonomous systems, and augmented as well as virtual reality, among others. 

\begin{figure}[t]
    \centering
    \includegraphics[width=\textwidth, trim={0cm 3cm 0cm 0cm}, clip]{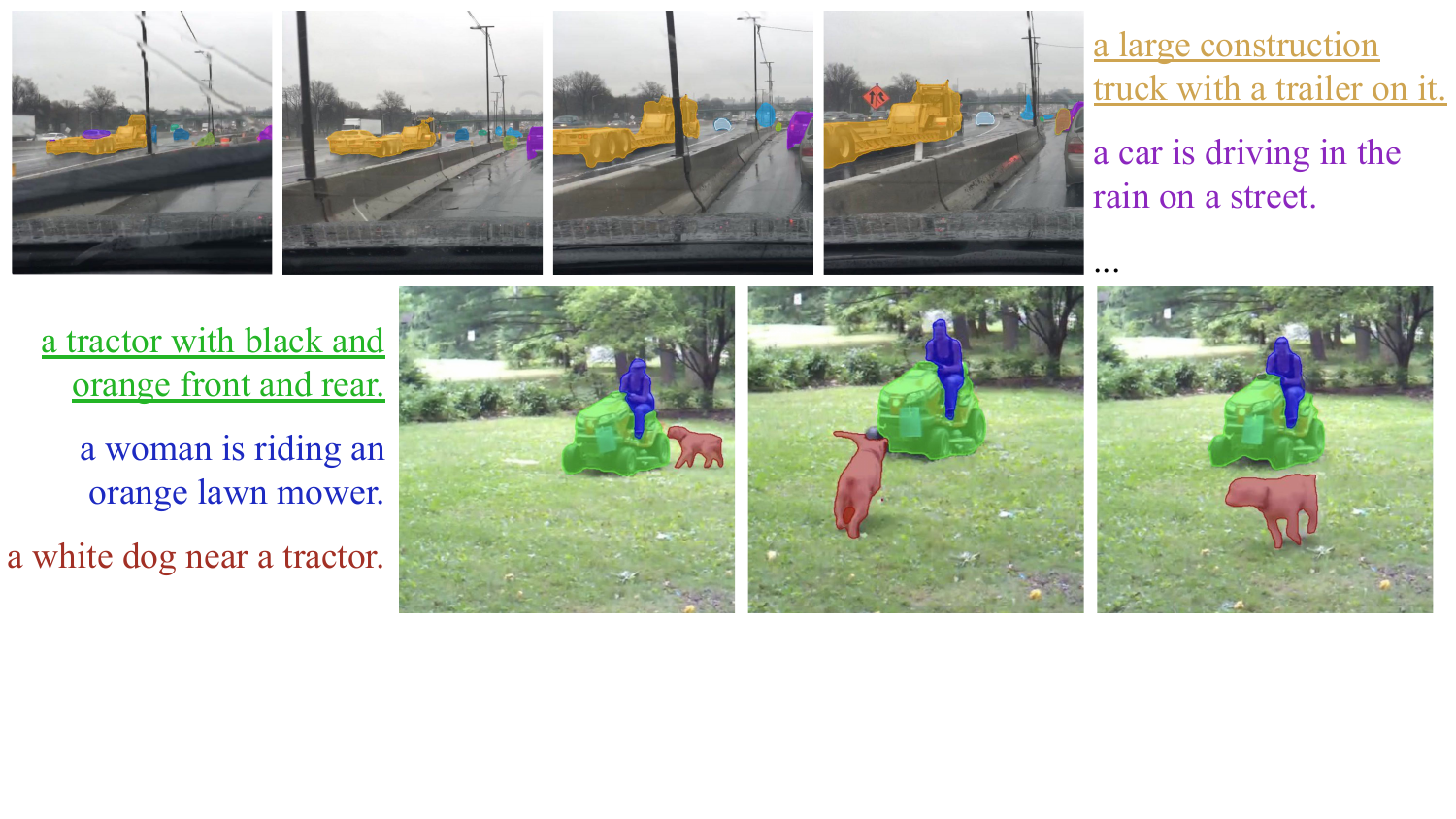}
  
    \caption{Our method, OW-VISCapTor, can simultaneously detect, segment, track, and caption objects in the given video frames. The first example (top row) shows a road scene with a previously unseen trailer truck and cars that are seen during training. The second example (bottom row) shows a person on a lawn mower and a dog on the grass. The lawn mower isn't part of the training set. We generate meaningful object-centric captions even for objects never seen during training. The captions for unseen objects are underlined.}
    \label{fig:qual_full}

\end{figure}

OW-VISCap generalizes open-world video instance segmentation (OW-VIS) which requires to deal with never-before-seen objects~\cite{tao, burst, uvo, owtb, owvisformer, deva}. For example, in Fig.~\ref{fig:qual_full}, the trailer truck (top row) highlighted in yellow, and the lawn mower (bottom row) highlighted in green, are never seen before during training. 
Current OW-VIS methods suffer from the following two main issues.

Firstly, all methods on video instance segmentation, closed- or open-world, assign a one-word label to the segmented objects. However, we argue that free-form captions are more descriptive than discrete class labels~\cite{capdet, grit}, and naturally facilitate zero-shot fine-grained captioning of objects, especially in an open-world setting. 
Notably, vision-language models have not been explored to unify open-world video instance segmentation and spatio-temporally dense fine-grained object captioning. 

Secondly, while classic OW-VIS methods~\cite{tao, burst, uvo, owtb, owvisformer, deva} rely on region-based object proposals~\cite{tao, burst, uvo, owtb}, more recent OW-VIS methods~\cite{owvisformer, deva} develop an ``abstractor'' to generate object queries. Abstractors~\cite{honeybee, blip2, instructblip, mask2former, detr} are used to extract and project important information from one space to another depending on the task, e.g., abstractors connect pixel and language spaces in vision and language models~\cite{blip2, instructblip}, pixel and object spaces in object detection/segmentation networks~\cite{mask2former, detr, owvisformer, deva}, etc. However, OW-VIS abstractors suffer from spatial information loss 
because they primarily focus on a few spatial regions, leading to a loss of finer spatial details~\cite{honeybee}. For closed-world object detection/segmentation networks, this can be compensated through extensive supervision~\cite{mask2formervideo, idol, minvis, caroq}, but it is challenging to address this issue for never-before-seen objects, i.e., when targeting an open-world setting. One way of overcoming this issue is to use a prompt as additional input from the user, ground truth or another network. The prompts can be in the form of points, bounding boxes, or text. However, these methods only work when the additional inputs are available, 
making them less practical in the real world. 

We study these issues in the newly proposed OW-VISCap task and propose the new baseline \textbf{OW-VISCapTor}, consisting of two   \textbf{O}pen-\textbf{W}orld \textbf{V}ideo \textbf{I}nstance \textbf{S}egmentation and \textbf{Cap}tioning abstrac\textbf{Tor}s. Note that OW-VISCap requires a more holistic object understanding: object representations need to be expressive and capture information not only for detecting and segmenting previously seen or unseen objects, but  also capture information for generating meaningful object-centric captions. This is an important step towards generalized scene understanding. OW-VISCapTor demonstrates this holistic object understanding by simultaneously detecting, segmenting and generating object-centric captions for objects in a video. Fig.~\ref{fig:qual_full} shows two examples in which our method successfully detects, segments, tracks and captions both closed- and open-world objects.

OW-VISCapTor addresses the first issue via an object-to-text abstractor which uses masked attention to project the object queries into text queries that can be interpreted by a frozen LLM to generate rich object-centric captions.
OW-VISCapTor addresses the second issue via an object abstractor that projects image features into  object queries. Spatial information is retained by finetuning a pretrained prompt encoder which is part of the object abstractor and  forms open-world object queries from points distributed evenly across video frames. 

Our approach improves upon a  baseline that simply combines prior work: 
open-world video instance segmentation (OW-VIS)~\cite{tao, burst, uvo, owtb, owvisformer, deva} and dense video captioning (Dense VOC)~\cite{densevoc2023}.  Also note, there are no existing datasets directly collected for the generalized OW-VISCap task. Yet, OW-VIS and Dense VOC cover all aspects of OW-VISCap: open-world object discovery, video instance segmentation and dense object-centric captioning. Compared to the baseline, we improve results by $13\%$ on the unseen categories for OW-VIS (BURST~\cite{burst} data), and by $10\%$ on the captioning accuracy for Dense VOC (VidSTG~\cite{vidstg} data). To further demonstrate the generalizability and efficacy of OW-VISCapTor, we compare our approach with the specialized state-of-the-art (SOTA) on OW-VIS and Dense VOC. Note that the OW-VIS SOTA can't be used for Dense VOC and vice-versa. Our generalized approach improves upon individual SOTA methods by $\sim6\%$ on the previously unseen (uncommon) categories in the BURST~\cite{burst} data, and by $\sim7\%$  on the captioning accuracy for detected objects on the VidSTG~\cite{vidstg} data. We also perform similar to the specialized state-of-the-art on the closed-world video instance segmentation task on the OVIS~\cite{ovis} data,  demonstrating generalizability.

\begin{figure}[t]
    \centering
    \includegraphics[width=\textwidth, trim={1.1cm 0cm 1.5cm 0cm}, clip]{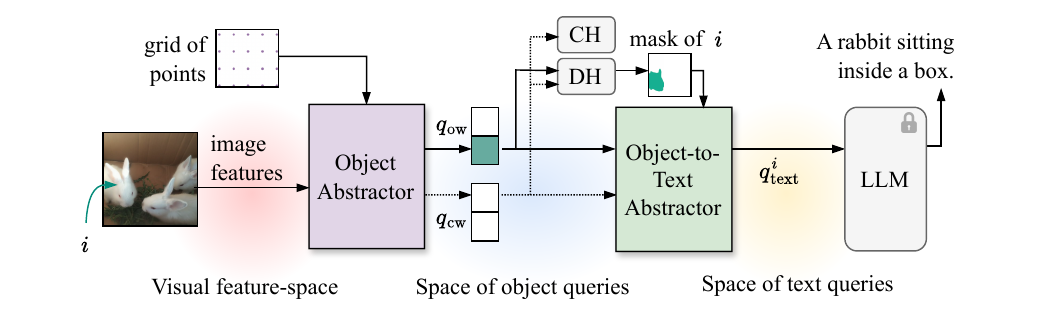}

    \caption{Overview of  OW-VISCapTor (Sec.~\ref{sec:app:overview}): an object abstractor (Sec.~\ref{sec:app:owqueries})  connects the image feature space to the object query space, 
    and an object-to-text abstractor (Sec.~\ref{sec:app:caphead})  connects the object query space to the text query space. 
    DH and CH stand for detection head and classification head.}

    \label{fig:figmain}

\end{figure}


\section{Related Work}

\label{sec:rel}

\subsection{Abstractors as Versatile Projectors}

\noindent \textbf{Abstractors in MLLMs.}
Abstractors have been immensely successful in Multimodal Large Language Models (MLLMs), connecting frozen vision encoders and a frozen LLM. BLIP-2~\cite{blip2} successfully adapted LLMs to visual tasks, showing notable zero-shot generalization and in-context learning capabilities. More recently, abstractors are used to enhance MLLMs through visual instruction tuning~\cite{instructblip, minigpt4, honeybee}. However, to our best knowledge, abstractors have not been explored for fine-grained captioning of objects in videos in the open-world, which is of interest here. 


\noindent \textbf{Abstractors for object discovery.}
Object abstractors can generate powerful object queries, useful for  closed-world object detection and segmentation  in both images~\cite{detr, maskformer, mask2former} and videos~\cite{mask2formervideo, idol, minvis, caroq}. More recently, abstractors have been used for open-world object discovery~\cite{owvisformer, deva}. However, they suffer from a spatial information loss because they primarily focus on a few spatial regions, leading to a loss of finer spatial details~\cite{honeybee}. This can be compensated through extensive supervision in the closed-world~\cite{mask2formervideo, idol, minvis, caroq}, but is challenging to address in the open-world for unseen objects. Prompt-based methods~\cite{sam, xdecoder, clipseg} can overcome this information loss, but use of prompts is often not realistic. In this work, we operate in a promptless open-world setting.

\subsection{Generalized Video Understanding Tasks}

Recently, there has been progress in unifying different video related tasks. TubeFormer~\cite{tubeformer}, Unicorn~\cite{unicorn}, and CAROQ~\cite{caroq} unify different video segmentation tasks in the closed world. DVOC-DS~\cite{densevoc2023} unifies the tasks of detecting (but not segmenting) closed-world objects in videos and captioning those closed-world objects. In this work, we explore the task of detecting and segmenting both closed- and open-world objects in videos, and captioning these objects.

Video understanding often starts from a strong generalized image understanding. 
Some methods~\cite{maskformer, mask2former, oneformer} unify different image segmentation methods, and provide a baseline for many different video understanding tasks~\cite{mask2formervideo, caroq, minvis, idol}.  X-Decoder~\cite{xdecoder} unifies different image segmentation tasks along with the task of referring image segmentation. SAM~\cite{sam} introduces a vision foundation model, that primarily performs prompt-based open-world image segmentation, and can be used for many downstream tasks. Different from these works, we develop a generalized method for videos that tackles segmentation and object-centric captioning for both  open- and closed-world objects.

\subsection{Specialized Video Understanding Tasks}

We briefly review OW-VIS, Dense VOC and VIS, and detail these tasks in Appendix~\ref{sec:append:rel}.

\textbf{Open-world video instance segmentation.} OW-VIS methods can be categorized into prompt-less methods~\cite{owtb,uvo,owvisformer, deva, burst} that either operate on classic region-based object proposals or suffer from spatial information loss; or promp-based methods that use prompts in the form of masks~\cite{vos, vostracking,vostracking2,videomatch,fastmatch,stm,stcn,feelvos,fastvos}, words~\cite{clipseg}, points~\cite{sam}, etc. We operate in a prompt-less setting, but spatially enrich an object-abstractor to generate open-world object queries.

\textbf{Dense video object captioning.}
DVOC-DS~\cite{densevoc2023} performs object-centric captioning of closed-world objects, but cannot caption multiple action segments or process long videos like many other video models~\cite{vivit, langasqueries, endtoendvis} .
Unlike DVOC-DS~\cite{densevoc2023}, we operate in the open-world, leverage masked attention for dense video object captioning and can address the aforementioned drawbacks. 

\textbf{Closed-world video instance segmentation.}
Methods for closed-world VIS either rely on classical region-based object proposals~\cite{vis, ovis, maskprop, stemseg, crossvis, asmots, neuralsolver, lift, motsfusion, trackrcnn, pointtrack}, or on abstractor-based object-queries~\cite{vistr, idol, minvis, caroq, mask2formervideo, trackformer, seqformer}. Differently, in this work, we explore abstractors to generate both closed- and open-world query-based proposals.

\begin{figure}[t]
    \centering
     \includegraphics[width=\textwidth, trim={1cm 0.2cm 1.1cm 0cm}, clip]{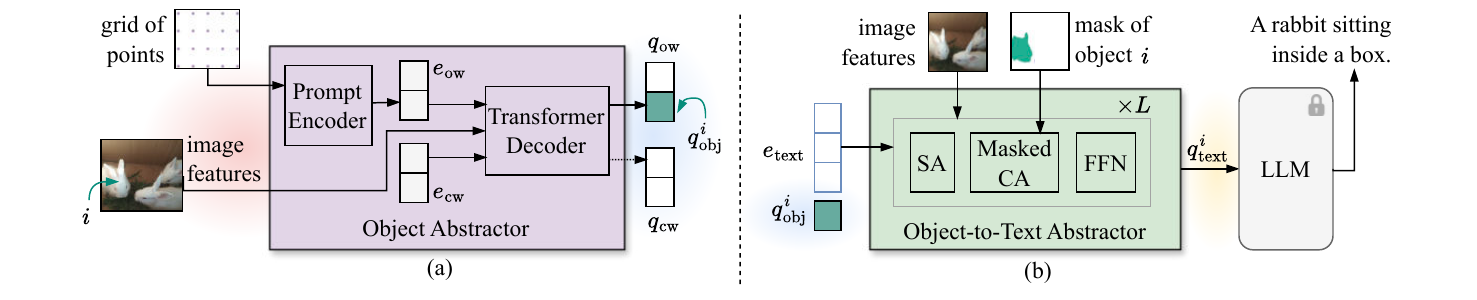}

    \caption{The proposed abstractors. (a) The \textbf{object abstractor} generates spatially rich open-world object queries $q_\mathrm{ow}$ from open-world embeddings $e_\mathrm{ow}$, and closed-world object queries $q_\mathrm{cw}$ from closed-world embeddings $e_\mathrm{cw}$. The open-world embeddings  $e_\mathrm{ow}$ are generated by encoding a grid of points 
    via a prompt encoder. 
    The closed-world embeddings are learnt. (b) The \textbf{object-to-text abstractor} 
    generates the object-centric text queries (e.g., $q^i_\mathrm{text}$ for the $i^\mathrm{th}$ object) that the frozen LLM uses for object-centric captioning. There are $L$ transformer blocks in the object-to-text abstractor, each one consisting of self-attention (SA), masked cross-attention (Masked CA), and a feed forward network (FFN).}
 
    \label{fig:abstractors}
\end{figure} 


\section{Abstractors for Open-World Video Instance Segmentation and Captioning}
\label{sec:app}

We propose to jointly address the tasks of 1) open-world video instance segmentation and 2) object-centric captioning: given a video,  we jointly detect, segment, track and caption object instances in a video. 
Importantly,  object instance categories may not be part of the training set (e.g., the trailer truck in Fig.~\ref{fig:qual_full} (top row)), placing our goal in an open-world setting.
To achieve this goal, we develop an approach which first breaks a given video into short clips, each consisting of $T$  frames. Each clip is processed using our  OW-VISCapTor. We discuss merging of the results of each clip in Appendix~\ref{sec:append:merge}.

We provide an overview of OW-VISCapTor 
in Sec.~\ref{sec:app:overview}. We then discuss our contributions: (a) the object abstractor that generates spatially-rich open-world object queries (Sec.~\ref{sec:app:owqueries}), along with our proposed inter-query contrastive loss, and (b) the object-to-text abstractor that uses masked cross-attention for object-centric captioning (Sec.~\ref{sec:app:caphead}). We discuss the final training objective in Sec.~\ref{sec:app:training}. 

\subsection{Overview}
\label{sec:app:overview}

Fig.~\ref{fig:figmain} provides an overview of our  OW-VISCapTor method. OW-VISCapTor consists of two abstractors: an object abstractor (Sec.~\ref{sec:app:owqueries} and Fig.~\ref{fig:abstractors} (a)) and an object-to-text abstractor (Sec.~\ref{sec:app:caphead} and  Fig.~\ref{fig:abstractors} (b)). The object abstractor connects the visual space of image features ($\mathbb{R}^{HWT \times C}$) to the space of object queries ($\mathbb{R}^{N_{\mathrm{obj}}\times C}$). Here, $N_\mathrm{obj}= N_\mathrm{obj, ow} + N_\mathrm{obj, cw}$ is the total number of object queries, which includes  the total number of open-world and closed-world queries,  $N_\mathrm{obj, ow}$ and $N_\mathrm{obj, cw}$. $C$ refers to the channel dimension for each object query and image feature. $H$, $W$ and $T$ refer to the height, width and clip-length. The object-to-text abstractor connects the space of object queries ($\mathbb{R}^{N_{\mathrm{obj}}\times C}$) and text queries ($\mathbb{R}^{{(N_\mathrm{text}+1)}\times C}$), to generate fine-grained object-centric captions. Here, $N_\mathrm{text}$ is the total number of text queries which are concatenated with individual object queries (hence the `$+1$') to generate text queries for each object.

To deal with the open-world setting, i.e., to ensure that we can detect, segment, track and caption never before seen object instances, the object abstractor generates spatially rich open-world object queries, $q_\mathrm{ow} \in \mathbb{R}^{N_{\mathrm{obj, ow}}\times C}$. These open-world object queries are in addition to closed-world object queries $q_\mathrm{cw} \in \mathbb{R}^{N_{\mathrm{obj, cw}}\times C}$, commonly used in prior work~\cite{mask2formervideo, mask2former, caroq, idol, minvis}. 
The open-world object queries are enriched spatially by a prompt encoder (shown in Fig.~\ref{fig:abstractors} (a)) that encodes a grid of points into prompt representations. Note,  by using open-world object queries, we  discover new and diverse open-world objects without needing additional prompts. 

We use $q_\mathrm{obj}=[q_\mathrm{ow}, q_\mathrm{cw}]$ to denote all the object queries obtained from the object abstractor. The $i^\mathrm{th}$ object query $q^i_\mathrm{obj} \in \mathbb{R}^{1\times C}$ is concatenated with learnt text embeddings $e_\mathrm{text}\in \mathbb{R}^{N_\mathrm{text}\times C}$. 
The concatenated object query and text embeddings are continuously modulated in the object-to-text abstractor by combining them with image features so as to generate meaningful text queries $q^i_\mathrm{text} \in \mathbb{R}^{(N_\mathrm{text}+1)\times C} $ for the $i^\mathrm{th}$ object. This is shown in Fig.~\ref{fig:abstractors} (b). Note, $q^i_\mathrm{text}$ is used by the frozen LLM to generate object-centric captions. The segmentation mask for the $i^\mathrm{th}$ object generated in the detection head is used to mask the attention in the object-to-text abstractor, as described in Sec.~\ref{sec:app:caphead}. We find this design to enable the LLM to generate more object-centric captions.

Both open- and closed-world object queries are processed by our object-to-text abstractor and LLM which yields an object-centric caption, and a detection head (DH in Fig.~\ref{fig:figmain}) which yields either a segmentation mask or a bounding-box. The closed-world object queries are further processed by a classification head (CH in Fig.~\ref{fig:figmain}) which yields a category label. 

\subsection{Object Abstractor}
\label{sec:app:owqueries}

The object abstractor is detailed in Fig.~\ref{fig:abstractors} (a). It consists of a prompt encoder, a transformer decoder and  closed-world trainable embeddings $e_\mathrm{cw}$.
The closed-world embeddings $e_\mathrm{cw}$ are modulated in the transformer decoder to generate the closed-world object queries $q_\mathrm{cw}$. We discuss the generation of open-world object queries next.

\noindent \textbf{Spatially-rich open-world object queries.} To help discover new objects, the object abstractor introduces open-world object queries $q_\mathrm{ow}$ in addition to the commonly used closed world object queries. 
Our open-world object queries are generated from open-world embeddings $e_\mathrm{ow}$, which are continuously modulated in the transformer decoder by combining them with image features via masked attention, following \cite{mask2former, mask2formervideo}. 

\noindent \textbf{Prompt encoder.} The open-world embeddings $e_\mathrm{ow}$ are generated in the prompt encoder. An illustration is provided  in Fig.~\ref{fig:abstractors} (a). We  encode a grid of equally spaced points along the height and width of the frames of the clip, using the prompt encoder employed in SAM~\cite{sam}. 
The use of equally spaced points encourages the open-world object queries to focus on different regions of the video frames, making them spatially rich and encouraging object discovery throughout the frames. This also encourages the open-world object queries to be diverse from one another.

\noindent \textbf{Inter-query contrastive loss.}
\label{sec:app:contloss}
We introduce an inter-query contrastive loss $\mathcal{L}_\mathrm{cont}$ to ensure that the object queries are different from each other. For closed-world objects, this loss helps in removing highly overlapping false positives. For open-world objects, it helps in the discovery of new objects.
Formally, 
\begin{align}
    \mathcal{L}_\mathrm{cont}=-\sum_{i,j}{L_1(q^i_\mathrm{obj}, q^j_\mathrm{obj})},
\end{align}
where  $q^i_\mathrm{obj}$ and $ q^j_\mathrm{obj}$ are the $i^\mathrm{th}$ and the $j^\mathrm{th}$ objects and $i \neq j$. $L_1$ refers to the L1 distance. Via this loss, we \emph{maximize} the L1 distance between the object queries, i.e., we encourage that the object queries differ from each other.

\subsection{Object-To-Text Abstractor} 
\label{sec:app:caphead}

The object-to-text abstractor (Fig.~\ref{fig:abstractors} (b)) connects the space of object queries and the space of text queries. 
It consists of $L$ transformer blocks, each block consisting of self-attention (SA), masked cross-attention (masked CA) and a feed forward network (FFN) as shown in Fig.~\ref{fig:abstractors} (b) to generate object-centric text queries. 
Next, we discuss  masked cross-attention.
 
\noindent \textbf{Masked cross-attention for object-centric captioning.} Masked cross-attention involves attending within the foreground region of the predicted mask for each object query. Concretely, to generate object-centric text queries $q^i_\mathrm{text}$ for each object $i$, we restrict the cross attention in the object-to-text abstractor by using the segmentation mask of the object generated by the detection head. 
Intuitively, this enables the model to focus on local object-centric features, which are sufficient to update the text queries. Importantly, note that context information from the video frames can be gathered through the self-attention layers. The proposed design hence doesn't take away any information. It rather provides the same information in clearly separated layers. 

Formally, for the $i^\mathrm{th}$ object we compute the  query features $X^{\mathrm{cap},i}_l \in \mathbb{R}^{(N_{\mathrm{text}}+1)\times C}$ obtained from the $l^\mathrm{th}$ object-to-text transformer layer via 
\begin{align}
        X^{\mathrm{cap},i}_l= \mathrm{softmax}(\mathcal{M}^i + Q^i_lK^T_l)V_l + X^{\mathrm{cap},i}_{l-1}.
    \label{eqn:cap_query}
    \end{align}
Here,  $\mathcal{M}^{i} \in \{0,-\infty \}^{1\times HWT}$ is the attention mask  in the object-to-text abstractor such that at feature location $(x,y)$, $\mathcal{M}^{i}(x,y)=0$, if $M^{i}(x,y)=1$ and $\mathcal{M}^{i}(x,y)=-\infty$, if $M^{i}(x,y)=0$. $M^{i}$ is the binary mask obtained from the detection head. Moreover, $K_l, V_l \in \mathbb{R}^{HWT \times C}$ are the linearly transformed image features. To initialize, we let $X^{\mathrm{cap},i}_0 =[q^i_\mathrm{obj},e_\mathrm{text}]$, where $q^i_\mathrm{obj} \in \mathbb{R}^{1\times C} $ is the $i^\mathrm{th}$ object query obtained from the object abstractor and $e_\mathrm{text} \in \mathbb{R}^{N_{\mathrm{text}}\times C}$ are the learnt text embeddings shown in Fig.~\ref{fig:figmain} and introduced in Sec.~\ref{sec:app:overview}. 


\subsection{Training}
\label{sec:app:training}

Our  training objective is $$\mathcal{L}_\mathrm{total}=\mathcal{L}_\mathrm{cont} + \mathcal{L}_\mathrm{cap} + \mathcal{L}_\mathrm{cw}+\mathcal{L}_\mathrm{ow}.$$
Here, $\mathcal{L}_\mathrm{cont}$ is the inter-query contrastive loss discussed in Sec.~\ref{sec:app:contloss}. $\mathcal{L}_\mathrm{cap}$ is the standard captioning loss introduced by DVOC-DS~\cite{densevoc2023}. $\mathcal{L}_\mathrm{cw}$ is the closed-world loss following prior work~\cite{mask2former}. To compute $\mathcal{L}_\mathrm{cw}$, the ground truth objects are first matched with the predicted closed-world objects; the optimal matching is used to compute the final closed-world loss $\mathcal{L}_\mathrm{cw}$. This loss consists of a detection/segmentation loss and a cross-entropy loss for predicting the closed-world object categories.
$\mathcal{L}_\mathrm{ow}$ is the open-world loss, which consists of only a detection/segmentation loss, unlike $\mathcal{L}_\mathrm{cw}$. The open-world loss is detailed in Appendix~\ref{sec:append:ow_loss}. 
Note, that the training data consists of only closed-world objects. We match the closed-world ground truth objects twice, once with the predicted open-world objects to compute $\mathcal{L}_\mathrm{ow}$, and once with the predicted closed-world objects to compute $\mathcal{L}_\mathrm{cw}$. We train the object abstractor, the object-to-text abstractor, and the image-feature extractor that generates the image-features as illustrated in  Fig.~\ref{fig:figmain}. The parameters of the LLM are frozen.

\begin{table}[t]
    \centering
      \setlength{\tabcolsep}{1.5pt}
    \footnotesize
     \caption{Results on OW-VIS (left) and Dense VOC (right). Onl.~refers to online frame-by-frame processing.  The columns highlighted in blue (OWTA for Unseen categories in OW-VIS and CapA for Dense VOC) highlight the `open-world' and `captioning' capabilities of different methods. The best scores are highlighted in bold font, and the second-best scores are underlined. 
    } 
    \begin{tabular}{c|c|acc|acccc}
    \toprule
& & \multicolumn{3}{c|}{\bf OW-VIS (OWTA)} & \multicolumn{5}{c}{\bf Dense VOC} \\ 
\hline
Method & Mode & Unseen & Overall & Seen  & CapA & CHOTA  & DetA & AssA & AP\textsubscript{M} \\ 
\hline \hline
OWTB~\cite{owtb} & onl. & 38.8 & 55.8 & 59.8 &- & -& -& -& -\\ 
Mask2Former~\cite{mask2former}+STCN~\cite{stcn} & onl. & 25.0 & 64.6 & 71.0 & -& -& -& -& -\\ 
Mask2Former~\cite{mask2former}+DEVA~\cite{deva} & onl. & 42.3 & \bf 69.5 & \bf 74.6 & -& -& -& -& -\\ 
EntitySeg~\cite{entityseg}+DEVA~\cite{deva} & onl. & 49.6 & 68.8 & 72.7 & -& -& -& -& - \\ 
\hline
DVOC-DS (joint training)~\cite{densevoc2023} & off. & -& -& -&  36.8& 51.6 & \bf 65.5 & \bf 56.9 & \bf 69.3 \\ 
DVOC-DS (disjoint training)~\cite{densevoc2023} & off. & - &-  & -& 10.0 & 28.0  & 45.9 & 48.0 & 39.8 \\ 

\hline \hline
Mask2Former~\cite{mask2former}+DEVA~\cite{deva}+ BLIP2~\cite{blip2} & onl. & 42.3 & \bf 69.5 & \bf 74.6 & 34.0 & 48.5 & 59.6 & \underline{56.4} & 60.1\\
\textit{OW-VISCapTor+CAROQ~\cite{caroq} (online)} & \textit{onl.}  & \textit{\underline{50.0} }&\textit{ 66.1 }&\textit{ 63.0} & \bf \textit{ 43.9}& \bf \textit{  53.1 } & \textit{\underline{60.1}} &\textit{ 54.0} & \textit{62.6} \\ 
\textit{OW-VISCapTor+DEVA~\cite{deva} (online)} & \textit{onl.} &  \bf \textit{55.2 } & \textit{\underline{69.0}} & \textit{\underline{73.5}}  & 
\textit{\underline{40.1}} & \textit{\underline{51.7}} &   \textit{60.0} & \textit{56.3} & \textit{ \underline{63.0}} \\ 

\bottomrule

\end{tabular}

    \label{tab:main}
\end{table}
    \begin{table}[t]
    \footnotesize
    \begin{minipage}{.42\linewidth}
   \centering
   \caption{Ablation on the BURST~\cite{burst} validation data. `w/o p.e.' refers to without prompt encoder; `w/o $\mathcal{L}_\mathrm{cont}$' refers to without contrastive loss.}

    \begin{tabular}{c|ccc} 
    \toprule
\bf Method & \multicolumn{3}{c}{\bf OWTA} \\
\midrule
 & Overall & Seen & Unseen \\ 
\hline
Ours & \bf 55.5 & \bf 58.2 & \bf 43.8 \\ 
w/o p.e.\ & 54.2 & 57.7 & 41.2 \\ 
w/o $\mathcal{L}_\mathrm{cont}$ & 53.5 & 56.1 & 41.6 \\
\bottomrule
\end{tabular} 
 
 \label{tab:abl_burst}
\end{minipage}
\hfill
    \begin{minipage}{.55\linewidth}
    \centering
    \caption{Ablation on the VidSTG~\cite{vidstg} data. `w/o m.a.' refers to without masked attention. `bb.~cap.' and `en.~bb.~cap.' refers to bounding box captioning and enlarged bounding box captioning.}

    \begin{tabular}{c|ccccc} 
    \toprule
\bf Method & \bf CHOTA & \bf DetA & \bf AssA & \bf CapA \\ 
\midrule
Ours & \bf 51.0 & 56.1 & 54.0 & \bf 43.9\\
w/o m.a. & 39.5&56.1 & 54.0& 20.3\\
bb.~cap. & 48.1 &56.1 & 54.0& 36.6\\
en.~bb.~cap. & 49.2& 56.1 &54.0& 39.4 \\
\bottomrule
\end{tabular} 

\label{tab:abl_vidstg}
\end{minipage}
\end{table}

\section{Experiments}
\label{sec:exp}
We evaluate OW-VISCapTor on the diverse tasks of open-world video instance segmentation (OW-VIS) and dense video object captioning (Dense VOC). Note that there is no dedicated dataset for our proposed task of open-world video instance segmentation \emph{and} captioning (OW-VISCap). Hence we use the two aforementioned tasks and evaluate the three different aspects of our approach: open-world capability, video instance segmentation and video object captioning.
In the following subsections, we first discuss the datasets and evaluation metrics used in our evaluation (Sec.~\ref{sec:exp:data_met}). We then compare our performance to a baseline that jointly addresses OW-VIS and Dense VOC, and also to specialized methods that address these two tasks individually (Sec.~\ref{sec:exp:main_res}). We demonstrate how our contributions result in better performance through an ablation study (Sec.~\ref{sec:exp:abl}). Finally, we show qualitative results (Sec.~\ref{sec:exp:qual}). In Appendix~\ref{sec:append:quant}, we also show results on closed-world VIS.

\begin{figure}[t]
    \centering
    \includegraphics[width=\textwidth, trim={0cm 3cm 0cm 0cm}, clip]{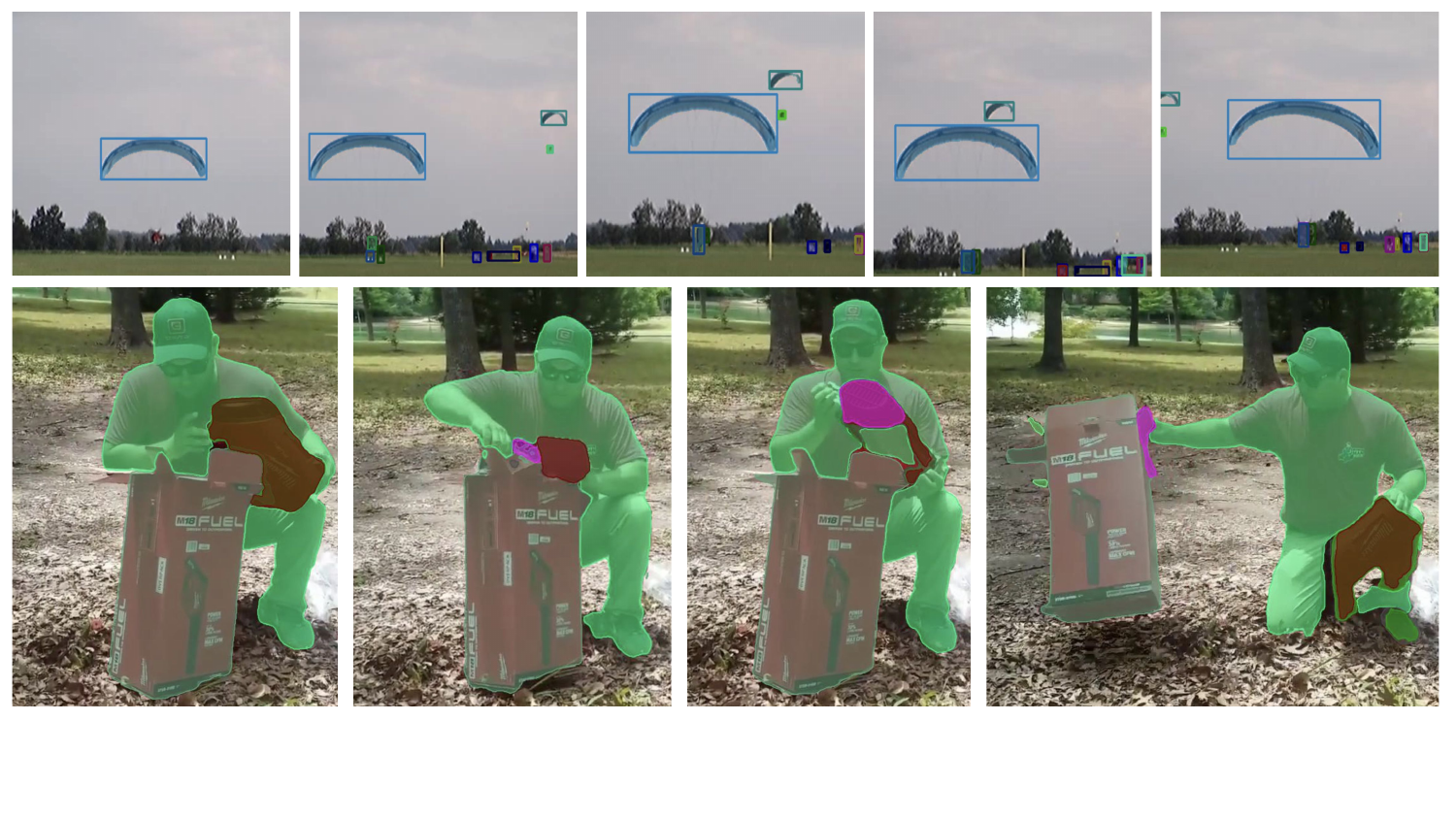}

    \caption{Example from the BURST validation data. The masks are superimposed on the objects. The top row shows examples of parachutes in the air and people on the grass. The parachutes belong to the uncommon object category, i.e., parachutes were never seen during training. Our approach detects and retains the identities of the blue and the green parachutes as the green parachute crosses the blue one. The bottom row shows a person unboxing a leaf blower. The carton of the leaf blower (gray mask), the leaf blower (maroon mask), and the plastic wrapper (pink mask) are never seen during training. We can consistently detect, segment, and track them along with the person (common object category during training).}

    \label{fig:qual_burst}
    
\end{figure}

\begin{figure}[t]
    \centering
    \includegraphics[width=\textwidth, trim={0cm 5cm 0.3cm 0cm}, clip]{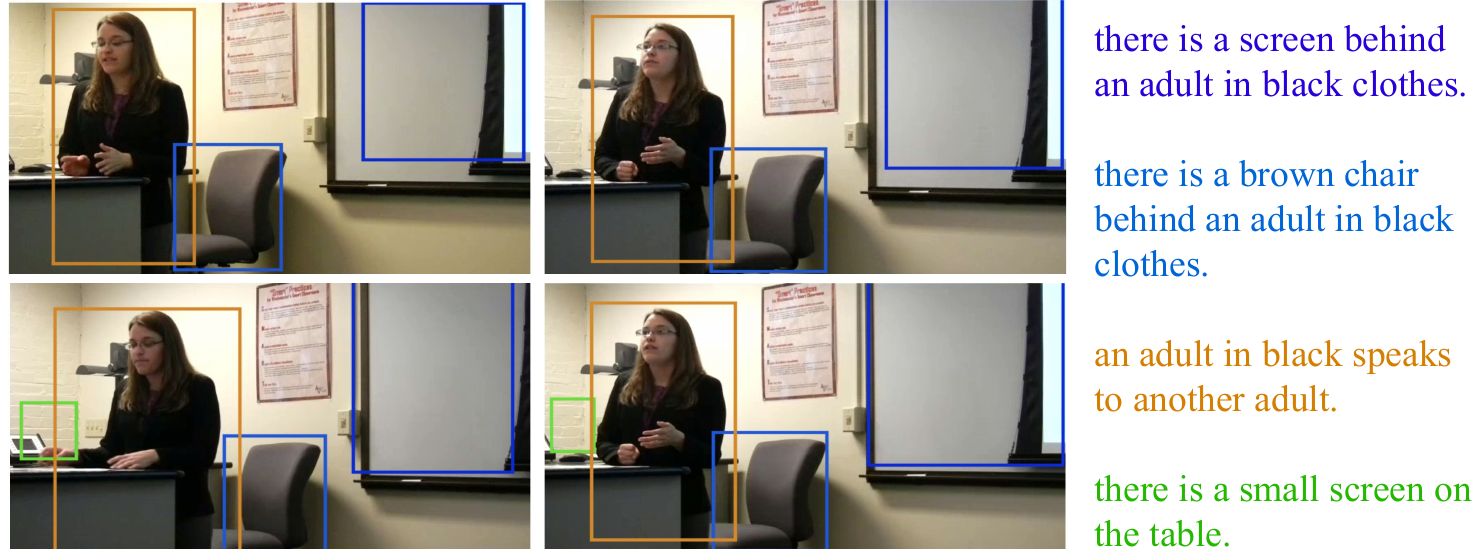}

    \caption{An example from the VidSTG data. Our approach is able to detect and track objects in the scene consistently and to generate meaningful object-centric captions for each of the detected objects.}

    \label{fig:qual_vidstg}

\end{figure}

\subsection{Datasets and Evaluation Metrics}
\label{sec:exp:data_met}

We evaluate our approach on the OW-VIS and Dense VOC tasks. For OW-VIS, we use the BURST dataset~\cite{burst}. For Dense VOC, we use the VidSTG dataset~\cite{vidstg}. Note, VidSTG~\cite{vidstg} has bounding box and tracking identity for all objects, but captions are not exhaustively provided. Following DVOC-DS~\cite{densevoc2023} 
the captioning loss for missing captions is removed during training and data with missing captions isn't evaluated at test time. 

For OW-VIS, we use the standard evaluation metrics of open-world tracking accuracy (OWTA)~\cite{burst} for all, common (seen) and uncommon (unseen) categories.
For Dense VOC, we use the captioned higher order tracking accuracy (CHOTA)~\cite{densevoc2023}, which depends on the detection accuracy (DetA), association accuracy (AssA), and captioning accuracy (CapA). We also report the frame-based METEOR score (AP\textsubscript{M}).

\subsection{Main Results}
\label{sec:exp:main_res}

We compare OW-VISCapTor with a generalized baseline, as well as specialized SOTA methods on the tasks of OW-VIS and Dense VOC. The results are summarized in Tab.~\ref{tab:main}. Even when compared to specialized SOTA on individual tasks, our method is able to achieve the best results (highlighted in bold) or the second-best results (underlined). Tab.~\ref{tab:main} (left) shows  results for the OW-VIS task on the BURST dataset~\cite{burst}. We report the open-world tracking accuracy for all, common (seen) and uncommon (unseen) categories. Tab.~\ref{tab:main}(right) shows  results on the Dense VOC task.

\noindent \textbf{Comparison with generalized baseline.}
We create a generalized baseline (Mask2Former~\cite{mask2former} + DEVA~\cite{deva} + BLIP2~\cite{blip2}) that is able to address the tasks of OW-VIS and Dense-VOC jointly. The generalized baseline consists of Mask2Former~\cite{mask2former} trained on the VidSTG~\cite{vidstg} dataset for object detection and integrated with DEVA~\cite{deva} for tracking. The per-frame predictions are first enlarged by $10\%$ to provide more overall image context. The enlarged bounding boxes are then used to crop the images for captioning using BLIP-2~\cite{blip2}. Note that Mask2Former~\cite{mask2former}+DEVA~\cite{deva} is a specialized previous SOTA on the OW-VIS task.

OW-VISCapTor outperforms Mask2Former~\cite{mask2former}+DEVA~\cite{deva}+BLIP2~\cite{blip2} consistently across both  datasets on unseen categories and object-captioning, demonstrating that our object queries are expressive enough to simultaneously detect, segment, track and caption seen or unseen objects. Mask2Former~\cite{mask2former}+DEVA~\cite{deva}+BLIP2~\cite{blip2}, in-spite of being SOTA on the previously seen object categories, struggles on the Dense VOC task. This is due to the lack of overall image-based context for captioning.

\noindent \textbf{Comparison with specialized SOTA.} OW-VISCapTor, integrated with DEVA~\cite{deva} for  temporal association of objects, achieves the state-of-the-art on the uncommon object categories (Tab.~\ref{tab:main} (left)), improving upon the next best method (EntitySeg~\cite{entityseg}+DEVA~\cite{deva}) by $\sim6$ points on the BURST~\cite{burst} validation data. 
 For the common categories, our method ranks $2^\mathrm{nd}$ in the BURST validation data. We use a SwinL~\cite{swin} backbone, and a clip-length of $T=1$ in this setting.

 Our method, when integrated with CAROQ~\cite{caroq} or DEVA~\cite{deva} for temporal association, outperforms DVOS-DS~\cite{densevoc2023} on the captioning accuracy (CapA), demonstrating that our object-to-text abstractor with masked cross-attention (Sec.~\ref{sec:app:caphead}) is effective in generating object-centric captions. We improve upon DS-VOC on the overall CHOTA metric, even though we slightly underperform on DetA and AssA. Note that DVOS-DS is an offline method: the entire object trajectories are used for generating the captions. Hence DVOS-DS  cannot process videos with more than $200$ frames. This is in contrast to our online method, where we sequentially process short video clips (of length $T=2$ for CAROQ~\cite{caroq} and of length $T=1$ for DEVA~\cite{deva}). 
DVOS-DS uses a ViT~\cite{vit} backbone, whereas we use SwinL~\cite{swin}, which leads to a difference in DetA scores.
We provide additional details on the merging of video clips 
in Appendix~\ref{sec:append:merge}.

Note that, even though the OW-VISCap task focuses on  generalizability, OW-VISCapTor outperforms specialized methods on the \textbf{open-world} metrics and the \textbf{captioning} metrics, demonstrating the effectiveness of our contributions: the object abstractor that processes our novel open-world object queries, and the novel object-to-text abstractor that generates rich object-centric captions.

\subsection{Ablation Studies}
\label{sec:exp:abl}

\noindent \textbf{Spatially-rich open-world object queries.}
Tab.~\ref{tab:abl_burst} (first and second row) shows that the spatially-rich open-world object queries $q_\mathrm{ow}$, described in Sec.~\ref{sec:app:owqueries}, help in discovering new objects. In Tab.~\ref{tab:abl_burst}, `w/o p.e.' refers to the setting without the prompt encoder  encoding spatial prompts 
into the open-world embeddings $e_\mathrm{ow}$. The open-world embeddings $e_\mathrm{ow}$ are trained like the closed-world embeddings $e_\mathrm{cw}$. 
We observe that the performance drops by $2.6$ points for uncommon categories compared to `Ours', even though the number of object queries are exactly the same in both settings. This highlights that the object abstractor suffers from spatial information loss if not augmented with a prompt encoder to encode spatial points.


\noindent \textbf{Contrastive loss.}
Tab.~\ref{tab:abl_burst} (first and last row) shows that the contrastive loss $\mathcal{L}_\mathrm{cont}$, described in Sec.~\ref{sec:app:contloss}, helps in detecting both the common (seen) and uncommon (unseen) categories of objects. The performance drops by $\sim2$ points for both the common and uncommon categories for the setting `w/o $\mathcal{L}_\mathrm{cont}$', i.e., when the contrastive loss is not used. The contrastive loss helps in removing highly overlapping false positives in the closed-world setting and in discovering new objects in the open-world setting. 

\noindent \textbf{Masked attention in object-to-text abstractor.}
Tab.~\ref{tab:abl_vidstg} shows that masked attention in the object-to-text abstractor, described in Sec.~\ref{sec:app:caphead}, helps in object-centric captioning. 
The second row `w/o m.a.' of Tab.~\ref{tab:abl_vidstg} refers to the setting without masked attention, i.e., the entire image-feature is used to calculate the cross-attention in the object-to-text abstractor. The object-centric context is only accumulated by concatenating the $i^\mathrm{th}$ object query with the learnt text embeddings, as discussed in Sec.~\ref{sec:app:caphead} and shown in Fig.~\ref{fig:figmain}. We observe that the captioning accuracy CapA drops by $23$ points, indicating that concatenating the object query with the text embeddings is not sufficient for an object-centric focus.
The third row in Tab.~\ref{tab:abl_vidstg}, `bb.~cap.' (bounding box captioning), pursues the opposite setting. Here, the images are cropped based on the object bounding box predictions in the detection head. The cropped images are directly used for captioning, ensuring that both the self- and cross-attention blocks in the object-to-text transformer operate on object-centric features. Note, that we don't use masked attention in this setting. We observe a drop in CapA of $5$ points. Although  cropping helps in retaining the object-centric information, the overall context from the entire image is missing. 
The fourth row in Tab.~\ref{tab:abl_vidstg}, `en.~bb.~cap.' (enlarged bounding box captioning), shows a similar setting as the third row, but the bounding boxes are first enlarged by $10\%$ to provide more overall image context. The enlarged bounding boxes are then used to crop the images for captioning. We observe a drop in CapA of $3$ points, indicating that enlarging the bounding boxes helps but is not sufficient to provide overall context. This is also highlighted in the third-last row and the last row in Tab.~\ref{tab:main}.



\subsection{Qualitative Results}
\label{sec:exp:qual}

Fig.~\ref{fig:qual_full} shows results on the BURST~\cite{burst} validation data.  OW-VISCap is able to simultaneously detect, segment, track and caption objects. The objects belong to both the open- and closed-world. Note that the BURST~\cite{burst} data doesn't provide object-centric captions for training, hence our object-to-text abstractor was not trained on BURST~\cite{burst} but only on VidSTG. 
We find this object-to-text abstractor to be effective in generating meaningful object-centric captions even for objects never seen during training. Fig.~\ref{fig:qual_burst} shows two examples from the BURST validation data. We can consistently detect, segment, and track previously seen and unseen objects. 
Fig.~\ref{fig:qual_vidstg} shows an example from the VidSTG~\cite{vidstg} data. Our method can detect, track and caption objects. Additionally, we discuss some failure modes of our method in Appendix~\ref{sec:append:lim}. 

\section{Conclusion}
\label{sec:conc}

We introduce OW-VISCapTor: two abstractors to \emph{jointly detect, segment, track, and caption previously seen or unseen objects in videos}. The developed object abstrator  generates spatially-rich open-world object queries which  encourage discovery of previously unseen objects without the need of additional user-input. 
Instead of assigning a fixed label to detected objects, we introduce an object-to-text abstractor that uses masked cross-attention to generate rich object-centric captions for each object.
\textbf{Societal Impact}. Our method can be used to segment and describe never before seen objects. 
This capability could be beneficial in assistive technologies for the blind, as well as in AR/VR. Although we don't see any direct ethical concerns, research in this direction makes video-processing technology increasingly accessible. This could encourage and increase malicious use and potentially create issues regarding unethical surveillance and privacy threats. This calls for stricter security measures both at a personal level (like password protecting video data), and societal level (like regulating open-source dataset and model releases).

\section{Acknowledgements}
This work is supported in party by the Agriculture and Food Research Initiative (AFRI) grant no.\ 2020-67021-32799/project accession no.\ 1024178 from the USDA National Institute of Food and Agriculture: NSF/USDA National AI Institute: AIFARMS. We also thank the Illinois Center for Digital Agriculture for seed funding for this project. Work is also supported in part by NSF under grants 2008387, 2045586, 2106825, MRI 1725729.


{\small
\bibliographystyle{unsrtnat}
\bibliography{neurips_2024}
}

\clearpage
\appendix
\appendix
\section*{\centering \Large Supplementary Material --- OW-VISCapTor: Abstractors for Open-World Video Instance Segmentation and Captioning}

\begin{figure}[ht]
    \centering
    \includegraphics[width=0.95\textwidth, trim={0cm, 3cm, 0cm, 0cm}, clip]{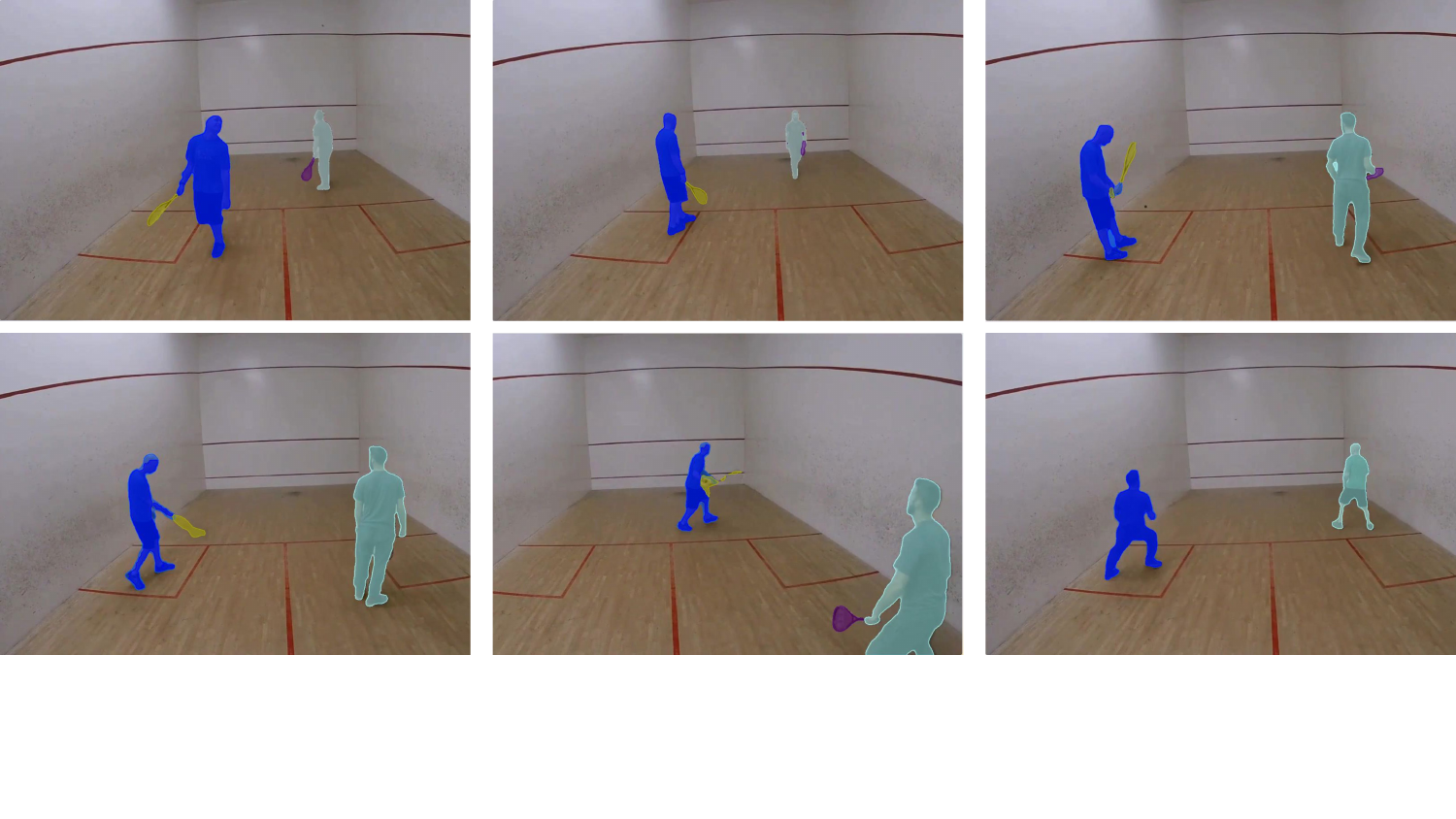}
    \caption{Results on the BURST dataset. We successfully segment and track closed-world objects (e.g., persons), and open-world objects (e.g., rackets) throughout the video.}
    \label{fig:qual_append_burst}

\end{figure} 

\begin{figure}[t]
    \centering
    \includegraphics[width=0.95\textwidth, trim={0cm, 6cm, 0cm, 0cm}, clip]{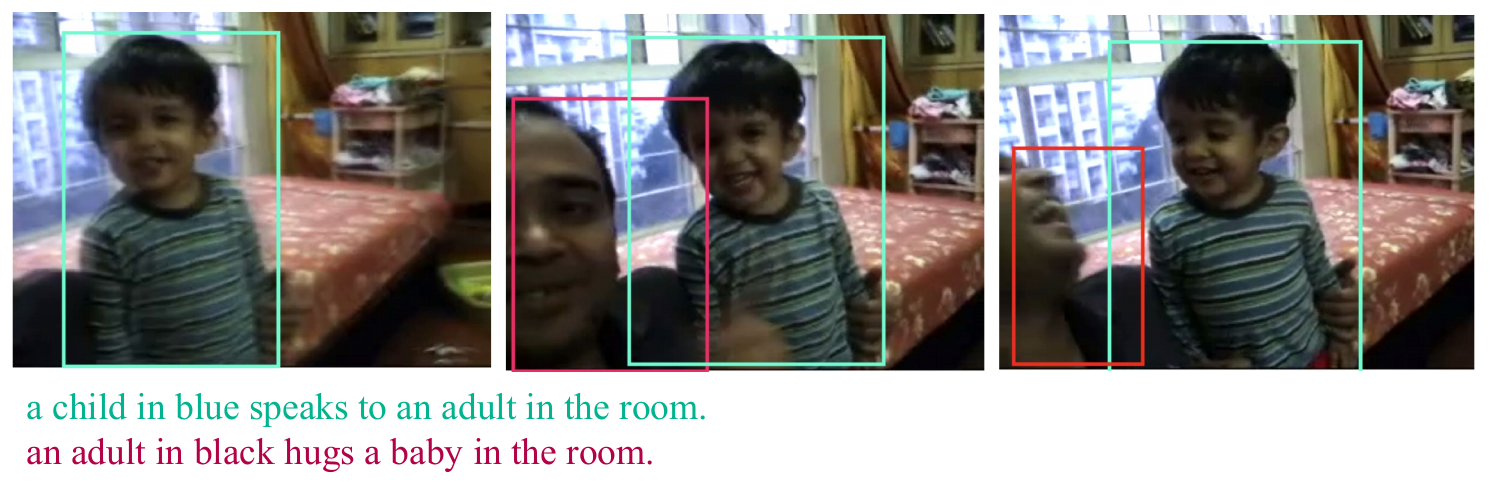}
    \caption{Results on the VidSTG dataset. Our approach detects, tracks and generates meaningful object-centric captions for each object throughout the video.}
    \label{fig:qual_append_vidstg}
\end{figure}

\noindent This is the supplementary material of the \textbf{OW-VISCapTor} paper, where we develop  \textbf{O}pen-\textbf{W}orld \textbf{V}ideo \textbf{I}nstance \textbf{S}egmentation and \textbf{Cap}tioning abstrac\textbf{Tor}s. We test our approach on the task of open-world video instance segmentation (OW-VIS) and dense video object captioning (Dense VOC) in Sec.~\ref{sec:exp} on the BURST~\cite{burst} and VidSTG~\cite{vidstg} datasets respectively. Fig.~\ref{fig:qual_append_burst} shows additional results on the BURST~\cite{burst} dataset. We successfully segment and track the closed-world objects: persons, and the open-world objects: rackets, throughout the video. Fig.~\ref{fig:qual_append_vidstg} shows results on the VidSTG~\cite{vidstg} dataset. The detected objects are tracked throughout the video and our method generates meaningful captions for each object.

In this supplementary material, we first discuss additional related works in Sec.~\ref{sec:append:rel}. In Sec.~\ref{sec:append:ow_loss}, we discuss how to train the open-world object queries introduced in Sec.~\ref{sec:app} of the main paper and then detail the merging of video-clips in Sec.~\ref{sec:append:merge}. We provide additional quantitative results (Sec.~\ref{sec:append:quant}) and qualitative analysis (Sec.~\ref{sec:append:analysis}) to support the contributions we made in Sec.~\ref{sec:app} of the main paper. Finally, we discuss the implementation details (Sec.~\ref{sec:append:implement}) and limitations (Sec.~\ref{sec:append:lim}) of our approach.

\section{Additional Related Work}
\label{sec:append:rel}

\subsection{Specialized Video Understanding Tasks}

We over-viewed some specialized video understanding tasks in Sec.~\ref{sec:rel} of the main paper. In this section, we provide additional details on these methods.

\textbf{Open-world video instance segmentation.} OW-VIS methods can be categorized into prompt-less and prompt-based methods.
Prompt-less methods~\cite{owtb,uvo,owvisformer, deva, burst} discover new objects based on an objectness score. Recent abstractor-based methods~\cite{owvisformer, deva} have shown to outperform classic methods using region-based object proposals~\cite{owtb, burst}.  However, these methods suffer from spatial information loss~\cite{honeybee}. Differently, we use an abstractor to generate spatially rich open-world object queries to address the OW-VIS task.
Prompt-based methods rely on prompts, i.e., prior knowledge,
to segment objects in videos. Prompts can be in the form of masks~\cite{vos, vostracking,vostracking2,videomatch,fastmatch,stm,stcn,feelvos,fastvos}, words~\cite{clipseg}, points~\cite{sam}, etc. Prompts provide a way to encode additional information to offset the spatial information loss observed with abstractors. However, in a true open-world setting, such prior knowledge may not be available. We operate in such a setting.


\textbf{Dense video object captioning.}
Dense video object captioning involves detecting, tracking, and captioning trajectories of closed-world objects in a video. For this task, DVOC-DS~\cite{densevoc2023} extends the image-based GRiT~\cite{grit} and trains it with a mixture of disjoint tasks. However, DVOC-DS~\cite{densevoc2023} cannot caption multiple action segments within a single object trajectory because the method produces a single caption for the entire object trajectory. In addition, similar to many other video models~\cite{vivit, langasqueries, endtoendvis}, DVOC-DS~\cite{densevoc2023} struggles with very long videos and only processes up to 200 frames. The method is further constrained to handle only known object categories and it is unclear how the method extends to an open-world setting.
Unlike DVOC-DS~\cite{densevoc2023}, we use abstractors to generate spatially rich open-world object queries and leverage masked attention for dense video object captioning. 
We also process video frames sequentially using a short temporal context and hence can process long videos, as well as handle multiple action segments within a single object trajectory. 

\textbf{Closed-world video instance segmentation.}
Closed-world video instance segmentation involves simultaneously segmenting and tracking objects from a fixed category set in a video. Some works~\cite{vis, ovis, maskprop, stemseg, crossvis, asmots, neuralsolver, lift, motsfusion, trackrcnn, pointtrack} rely on classical region-based proposals. Recent works~\cite{vistr, idol, minvis, caroq, mask2formervideo, trackformer, seqformer} rely on abstractor-based object queries and perform significantly better at discovering closed-world objects. Differently, in this work, we explore abstractors to generate query-based proposals for the closed- and open-world setting.

\subsection{Contrastive Loss for Object Queries}

Contrastive losses have been used to help in video instance segmentation. OWVISFormer~\cite{owvisformer} uses a contrastive loss in the open-world setting: it ensures that assigned foreground objects are similar to each other while being different from the background objects. IDOL~\cite{idol} works in the closed-world setting and uses an inter-frame contrastive loss to ensure object queries belonging to the same object across frames are similar, and object queries of different instances across frames differ. In contrast, in this work, for both the closed- and open-world setting, we use a contrastive loss to ensure that no two object queries in the foreground are similar to each other, even in the same frame.

\section{Open-World Loss} 
\label{sec:append:ow_loss}
To encourage the discovery of new objects, we introduce an open-world loss $\mathcal{L}_\mathrm{ow}$ in Sec.~\ref{sec:app:training} of the main paper. It differs in one key aspect from a classic closed-world loss \cite{mask2formervideo}: we don't have a cross-entropy loss. 

We first match the ground truth objects
with the open-world predictions by minimizing a
matching cost using the Hungarian algorithm~\cite{hungarian}. The optimal matching is then used to calculate the final open-world loss.
Let us use $\hat{\sigma}_\mathrm{ow}$  to represent the optimal matching between the ground truth objects and the open-world predictions.

Formally, the open-world loss is given by 
\begin{align}
    \mathcal{L}_\mathrm{ow}=\sum_i{
   \mathcal{L}_\mathrm{det}(i,\hat{\sigma}_\mathrm{ow}^i)},
\end{align}
where $i$ is the object index. 
$\mathcal{L}_\mathrm{det}(i,\hat{\sigma}_\mathrm{ow}^i)$ is the detection loss (mask loss or bounding box loss) between the ground truth object with index $i$ and a prediction with index $\hat{\sigma}_\mathrm{ow}^i$. 

Differently, in the classic closed-world setting, the object categories are available, which enables us to train a classification head (CH in Fig.~\ref{fig:figmain}) using a cross-entropy loss. 

\section{Merging of Clips}
\label{sec:append:merge}

To achieve open-world video instance segmentation and captioning, OW-VISCapTor first breaks a given video into short clips, each consisting of $T$  frames. Each clip is processed using our  OW-VISCapTor. We now discuss how the predictions of individual clips are merged.

We use DEVA~\cite{deva}, a recent state-of-the-art for temporal association, to connect video clips temporally, as highlighted in Tab.~\ref{tab:main} of the main paper. DEVA~\cite{deva} develops a class-agnostic temporal propagation approach to track detected or segmented objects. However, DEVA~\cite{deva} permits clip-length of $T=1$. To incorporate more temporal context, we also adopt CAROQ~\cite{caroq}, where the object queries are propagated again and again for subsequent video clips, thereby carrying temporal information forward, and implicitly tracking the objects. CAROQ~\cite{caroq} can operate with a clip-length of $T\geq1$. The effect of clip-length while using CAROQ~\cite{caroq} is ablated for the Dense VOC task in Tab.~\ref{tab:abl:cliplength}.

\begin{table}[t]
    \centering
    \footnotesize
    \setlength{\tabcolsep}{3pt}
    \caption{Open-world tracking accuracy (OWTA) on the BURST test dataset for uncommon (unseen), overall and common (seen) categories of objects. The best scores are highlighted in bold font, and the second-best scores are underlined.
    }

    \begin{tabular}{c|acc} 
    \toprule
\bf Method  & \multicolumn{3}{c}{ \bf Test (OWTA)} \\ 
\midrule
  & Unseen & Overall & Seen  \\ 
\hline
OWTB~\cite{owtb} & 38.3 &  56.0 & 59.9  \\ 
Mask2Former~\cite{mask2former}+STCN~\cite{stcn}  & 23.9& 57.5 & 62.9  \\ 
Mask2Former~\cite{mask2former}+DEVA~\cite{deva}  & 44.1& \bf 70.1 & \bf 75.0  \\ 
EntitySeg~\cite{entityseg}+DEVA~\cite{deva}  & \underline{53.0}&  \underline{69.5} & \underline{72.9}  \\ 
OW-VISCapTor (ours) + DEVA~\cite{deva} & \bf 57.2& 69.2 & 72.4 \\ 
\bottomrule
\end{tabular} 
    
    \label{tab:burst}
\end{table}
\begin{table*}[t]
    \centering
    \footnotesize
    \setlength{\tabcolsep}{2.9pt}
     \caption{Results on the OVIS~\cite{ovis} validation data. All methods use the ResNet-50 backbone. The best performing methods are highlighted in bold font. The second best methods are underlined.}
\begin{tabular}{c|ccccc} 
\toprule
\bf Method & \bf AP & \bf AP\textsubscript{50} & \bf AP\textsubscript{75} & \bf AR\textsubscript{1} & \bf AR\textsubscript{10} \\ 
\midrule
MaskTrack ~\cite{vis} & 10.8 & 25.3 & 8.5 & 7.9 & 14.9 \\ 
DeVIS~\cite{devis} & 23.7 & 47.6 & 20.8 & 12.0 & 28.9 \\ 
MinVIS~\cite{minvis} & 25.0 & 45.5 & \underline{24.0} & 13.9 & 29.7 \\ 
VMT~\cite{vmt} & 16.9 & 36.4 & 13.7 & 10.4 & 22.7 \\ 
InstanceFormer~\cite{instanceformer} & 20.0 & 40.7 & 18.1 & 12.0 & 27.1 \\ 
VITA~\cite{vita} & 19.6 & 41.2 & 17.4 & 11.7 & 26.0 \\ 
CAROQ~\cite{caroq} & \bf 25.8 & \underline{47.9} & \bf 25.4 & \underline{14.2} & \bf 33.9 \\ 
OW-VISCapTor (w/o $\mathcal{L}_\mathrm{cont}$)& 23.2& 45.2&21.7 & 13.5&30.1 \\ 
OW-VISCapTor (ours) & \underline{25.4}& \bf 48.8 & 22.8& \bf 14.3 & \underline{32.8}\\ 
\bottomrule
\end{tabular}

    \label{tab:ovis}
\end{table*}
\begin{table}[t]
     \caption{Ablation to show how the clip-length $T$ affects the performance on the DenseVOC task.}
\footnotesize
    \centering
    \setlength{\tabcolsep}{3pt}
\begin{tabular}{c|ccccc}
\toprule
\bf T & \bf CHOTA & \bf CapA & \bf DetA & \bf AssA\\ 
\midrule
1 & 49.2&  40.5 & 58.8 & 50.0\\
2 & \bf 53.0 & \bf 43.9 & \bf 60.1 & 54.0\\
4 & 52.3 &  43.6&  59.6& \bf 55.1 \\
\bottomrule
\end{tabular}  
    \label{tab:abl:cliplength}
\end{table}
 
\section{Additional Quantitative Results}
\label{sec:append:quant}

\textbf{Results on BURST test data.}
Tab.~\ref{tab:burst} shows the results of our method on the BURST test data. Our method performs the best on unseen categories, which is consistent with the results  on the BURST~\cite{burst} validation data shown in Sec.~\ref{sec:exp:main_res}.

\textbf{Results on video instance segmentation (OVIS data). }
Tab.~\ref{tab:ovis} shows our results for the closed-world video instance segmentation task on the OVIS~\cite{ovis} dataset. In the closed-world setting, we disable the open-world object queries. We notice that the contrastive loss $\mathcal{L}_\mathrm{cont}$ (discussed in Sec.~\ref{sec:app:contloss}) improves the closed-world results. We use a clip-length of $T=2$ in this setting and  CAROQ~\cite{caroq} to combine results from video clips.

\textbf{Effect of clip-length $T$.} Tab.~\ref{tab:abl:cliplength} shows how length $T$ of the video-clips affects performance, when CAROQ~\cite{caroq} is used to merge the video-clips. $T=1$ defaults to frame-by-frame inference. We clearly observe that frame-by-frame inference is sub-optimal as compared to a larger temporal context.
However, the performance improvement is marginal with $T>2$. This suggests that the object queries already retain long-term temporal information and additional context features are no longer significant, as also highlighted in CAROQ~\cite{caroq}.

\textbf{Reproducibility.}
Sec.~\ref{sec:exp:main_res} discusses the main results on the BURST~\cite{burst} and VidSTG~\cite{vidstg} data.  Results are calculated by repeating the respective experiments $3$ times with different seeds. The variances in performance for OW-VISCapTor+DEVA and  OW-VISCapTor+CAROQ are $0.4$  on the unseen categories for the BURST~\cite{burst} dataset and $0.5$ on the captioning accuracy for the VidSTG dataset~\cite{vidstg} respectively. This shows that the proposed method produces consistent results despite different seeds.

\section{Additional Qualitative Analysis} 
\label{sec:append:analysis}
In this section, we provide additional qualitative analysis of the different components discussed in Sec.~\ref{sec:app} of the main paper.

\textbf{Open-world embeddings as spatially-rich object proposals.} The object abstractor in our approach introduces spatially rich open-world embeddings $e_\mathrm{ow}$ (Sec.~\ref{sec:app:owqueries} of the main paper). These embeddings are modulated in the transformer decoder (Fig.~\ref{fig:abstractors} (a)) to generate open-world object queries. We obtain the open-world embeddings by encoding a grid of equally spaced points across the feature dimensions through a prompt encoder. This encourages object discovery throughout the video frame. The open-world embeddings act as initial abstract object proposals. 

\begin{figure}[t]
    \centering
    \includegraphics[width=\textwidth]{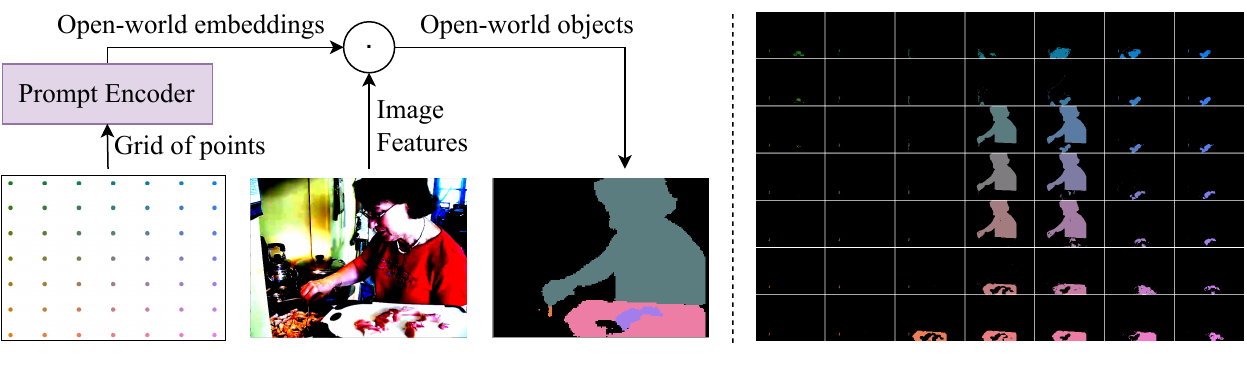}
    \caption{The prompt encoder in our proposed object abstractor creates spatially-rich open-world embeddings that act as strong object proposals. A strong spatial correlation exists between the grid of points and the segmentation masks generated by the corresponding open-world embeddings. This is highlighted by the color of the points in the grid and the color of the segmentation masks.}
    \label{fig:append_ow}
\end{figure}

In Fig.~\ref{fig:append_ow}, we show that the open-world embeddings are strong object proposals, even before they are modulated by the object transformer by being combined with video frame features. The person, the spoon, the plate, and some food on the plate are discovered by the open-world embeddings. Their segmentation masks are obtained by the dot product between the open-world embeddings and the video frame features. 
Further, we see a strong spatial correlation between the grid of points and the segmentation masks generated by the corresponding open-world embeddings. This suggests that encoding a grid of points using the prompt encoder makes our object abstractor spatially rich. 

\begin{figure}[t]
    \centering
    \includegraphics[width=\textwidth, trim={0cm, 5.5cm, 0cm, 0cm}, clip]{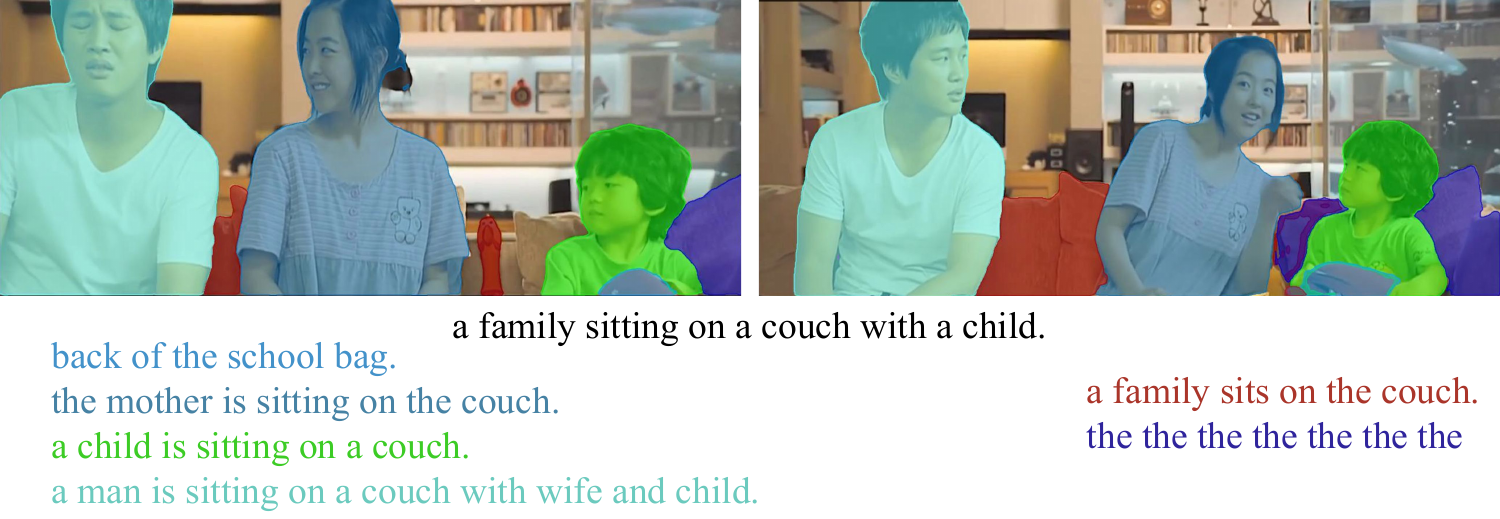}
    \caption{Masked cross-attention in the object-to-text abstractor helps generate object-centric captions (colored captions on the left). Providing no mask during cross attention fails to capture object-centric details (black caption). However, masked attention sometimes fails to generate meaningful object-centric captions for small objects (colored captions on the right).}
    \label{fig:append_masked_caption}
\end{figure}

\textbf{Masked cross-attention for object-centric captioning.}
In Tab.~\ref{tab:abl_vidstg}, we quantitatively show that masked cross-attention in the object-to-text abstractor (Sec.~\ref{sec:app:caphead}) helps in generating accurate object-centric captions for individual objects.  Fig.~\ref{fig:append_masked_caption} shows the high quality of the object-centric captions generated. The black colored caption (`a family sitting on a couch with a child') is obtained when no mask is provided in the object-to-text transformer. The entire image features are seen during the cross-attention operation in each layer of the object-to-text transformer. The caption fails to capture the object-centric details. 

Other colored captions are generated with masked attention for the individual objects. The colored captions on the left highlight the effectiveness of masked attention in generating object-centric captions. For example, the three persons (highlighted in cyan, grayish blue, and green) have distinct captions, each one describing the individual identities of the corresponding person. The school bag (light blue) is also described correctly. We want to note that sometimes the method fails to generate meaningful object-centric captions for small objects (captions on the right). We discuss this more in Appendix~\ref{sec:append:lim}.

\begin{figure}[t]
    \centering
    \includegraphics[width=\textwidth, trim={0cm, 6cm, 0cm, 0cm}, clip]{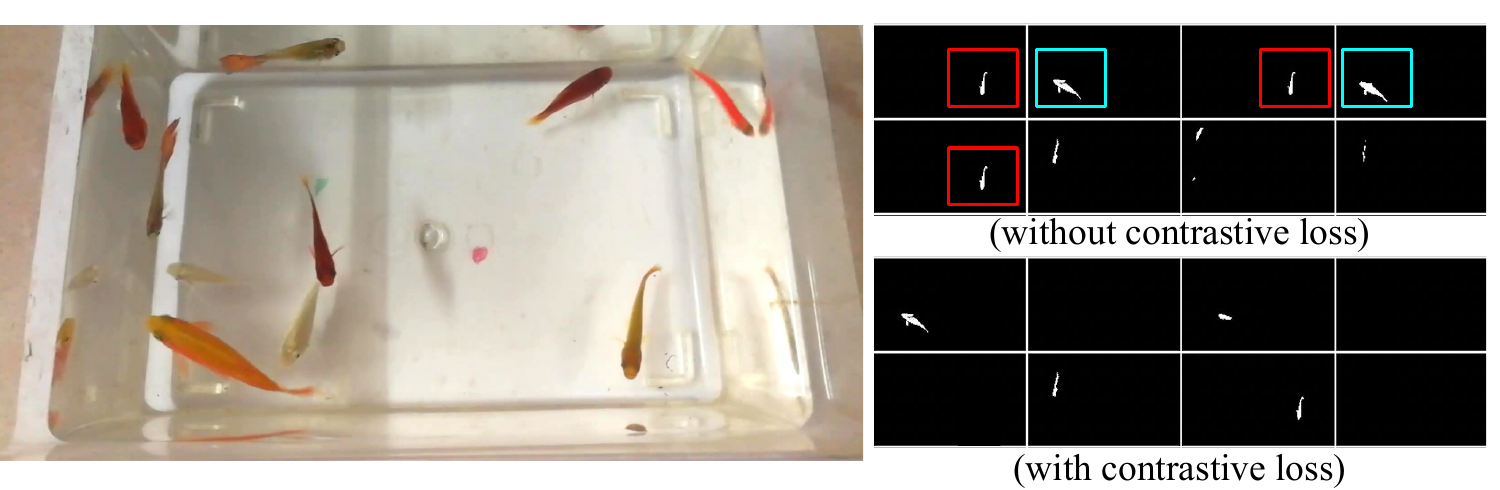}
    \caption{The repetitive predictions without contrastive loss (top-right) are highlighted with red and cyan boxes. Contrastive loss  (bottom-right) helps in suppressing these repetitions.}
    \label{fig:qual_append_contrastive}
\end{figure}

\begin{figure}[t]
    \centering
    \includegraphics[width=\textwidth, trim={0cm, 8cm, 0cm, 0cm}, clip]{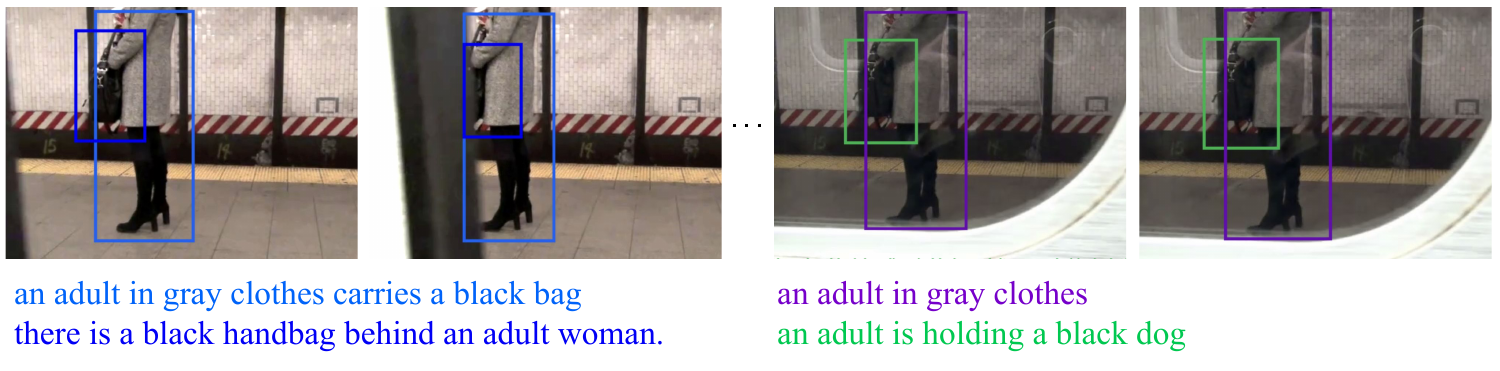}
    \caption{A failure mode, where the object identities aren't retained after prolonged occlusion. After a train crosses the screen for a prolonged period ($\sim30$ frames), the person initially identified as light blue is later identified as purple. The bag initially identified as dark blue is later confused to be a dog identified as green.}
    \label{fig:limitation}
\end{figure}

\textbf{Contrastive loss to suppress overlapping predictions.}
In Tab.~\ref{tab:abl_burst} of the main paper and in Tab.~\ref{tab:ovis}, we demonstrate the effectiveness of using the contrastive loss $\mathcal{L}_\mathrm{cont}$ (discussed in Sec.~\ref{sec:app:contloss}) for object detection. This loss encourages that object queries differ from each other, among others by suppressing highly overlapping predictions. We highlight this in Fig.~\ref{fig:qual_append_contrastive}. The left image shows a frame from the OVIS~\cite{ovis} dataset. The top-right and bottom-right images show a few predictions from our network trained without (top) and trained with (bottom) the contrastive loss. The repetitive predictions for the top-right image are highlighted with red and cyan boxes. The contrastive loss helps in removing these repetitions.


\section{Implementation Details}
\label{sec:append:implement}
In this section, we provide the implementation details of OW-VISCapTor. We first describe the architecture of different components discussed in Sec.~\ref{sec:app}. We then discuss our choice of hyper-parameters for all experiments discussed in Sec.~\ref{sec:exp}. We also discuss the resources and the licenses of the code bases and datasets used in the paper.

\subsection{Architecture} 

\noindent \textbf{Object abstractor.} Our object abstractor consists of a prompt encoder, a transformer decoder, and closed-world embeddings $e_\mathrm{ow}$. Our prompt encoder, discussed in Sec.~\ref{sec:app:owqueries}, is lightweight and consists of $4$ learned embeddings with $256$ channels added to their positional encodings~\cite{fourier}.
We initialize our prompt encoder from SAM~\cite{sam} and fine-tune it on the BURST~\cite{burst} dataset to generate open-world embeddings. The transformer decoder has $3$ transformer layers, each layer consisting of a masked cross-attention, self-attention and FFN layer followed by a LayerNorm. The transformer layers and the closed-world embeddings  are initialized from Mask2Former trained on COCO instance segmentation, and fine-tuned on the BURST~\cite{burst} or VidSTG~\cite{vidstg} datasets. An auxiliary loss is added to every intermediate transformer decoder layer and to the closed-world embeddings following Mask2Former~\cite{mask2former} to improve the detection performance of OW-VISCapTor.

\noindent \textbf{Object-to-text abstractor.} Our object-to-text abstractor, detailed in Sec.~\ref{sec:app:caphead}, consists of $11$ transformer layers and the text embeddings $e_\mathrm{text}$, which are initialized from BLIP-2~\cite{blip2}. 
Each transformer layer consists of a self-attention, and a FFN layer, followed by a LayerNorm. The masked cross-attention is present in alternate layers after the self-attention, following BLIP-2~\cite{blip2}. The object queries (with channel dim.\ $256$) are first passed through a linear layer to match the channel dimensions of the text embeddings (with channel dim.\ $768$). Each of these modified object queries is then concatenated with the text embeddings and modulated in the object-to-text transformer to generate object-centric text queries. We fine-tune the object-to-text abstractor, the linear layer, and the text embeddings on the VidSTG~\cite{vidstg} dataset. The object-centric text queries are used by an LLM for object-centric captioning. We use a frozen OPT-2.7B model as the LLM, which is a decoder-only model having 2.7 billion parameters.  

\noindent \textbf{Feature extractor.} The feature extractor for each video frame, consisting of a backbone and a pixel-decoder, takes the video frame as input and produces multi-level image features. This design follows the meta-architecture from Mask2Former~\cite{mask2former}. The transformer decoder in the object abstractor modulates the open- and closed-world object queries using the multi-level image features. We use the ResNet-50 and the SwinL backbones for our experiments. The Deformable-DETR~\cite{deformabledetr} is used as the pixel decoder, with $6$ multi-scale deformable attention layers applied to feature maps to generate multi-scale image features of resolution $1/4$, $1/8$, $1/16$ and $1/32$ with $256$ channels. These image features are used as input in the transformer decoder of the object transformer in a round robin fashion following Mask2Former~\cite{mask2former}. Note the image features which act as inputs to the object-to-text abstractor are obtained from the vision encoder of BLIP-2~\cite{blip2}.

\noindent \textbf{Detection head and classification head.} The detection head consists of $2$ linear layers separated by ReLU activation layers. It generates either bounding box predictions (VidSTG~\cite{vidstg} dataset) or predictions that yield the segmentation masks (BURST~\cite{burst} dataset) when a dot product is computed between them and the image-features. 
The classification head for the closed-world objects consists of a linear layer to generate the logits for each category.

\subsection{Hyper-Parameters}
We now discuss the hyper-parameters used in this work.
 For experiments on the BURST~\cite{burst} dataset, we encode a grid of $7\times7$ points across the width and height of the image features to obtain the open-world embeddings $e_\mathrm{ow}$ discussed in Sec.~\ref{sec:app:owqueries}. Hence the total number of open-world object queries $N_\mathrm{obj, ow}$ is $49$. We also experimented with a grid of $4\times 4$ and $10 \times 10$ but didn't see a significant change in performance. For experiments on the VidSTG~\cite{vidstg} dataset, the number of text embeddings $e_\mathrm{text}$  is $32$. In all experiments, the maximum number of closed world objects ($N_\mathrm{obj, cw}$) in a given video for a ResNet-50 backbone is $100$, and for a Swin-L backbone is $200$. We use a feature dimension $C$ (Sec.~\ref{sec:app:overview}) of $256$ in all models, unless stated otherwise.
 
 We trained the models with an initial learning rate of $0.0001$ and ADAMW~\cite{adamw} optimizer with a weight decay of $0.05$. We use a batch size of $8$. The backbone, the pixel decoder, the closed-world embeddings and the transformer layers in the object abstractor were first initialized with weights from Mask2Former~\cite{mask2former} trained on the COCO image instance segmentation dataset. The object-to-text abstractor and the text embeddings were first initialized with weights from BLIP-2~\cite{blip2}. 
 We then fine-tune the models on the respective BURST~\cite{burst}, VidSTG~\cite{vidstg}, and OVIS~\cite{ovis} datasets for $10,000$, $16,000$, and $8,000$ iterations respectively. 

\subsection{Resources}
We used $8$ NVIDIA A100 GPUs to run the experiments presented in this paper. Each experiment took roughly $10$ GPU hours of training on the A100 GPUs for the BURST experiments,  $16$ GPU hours for the VidSTG experiments, and $8$ GPU hours for the OVIS dataset experiments.

\subsection{Licenses}
Our code is built on Mask2Former \cite{mask2former} which is majorly licensed under the MIT license, with some portions under the Apache-2.0 License. We also build on SAM~\cite{sam}, which is released under the Apache 2.0 License, and BLIP-2~\cite{blip2} which is released under the MIT license. The VidSTG~\cite{vidstg} and BURST~\cite{burst} datasets are released under the MIT license. The OVIS~\cite{ovis} dataset is released under the Attribution-NonCommercial-ShareAlike (CC BY-NC-SA) License.

\section{Limitations}
\label{sec:append:lim}
In this section, we show some failure modes of our proposed approach and discuss limitations. 
Fig.~\ref{fig:append_ow} shows that our approach sometimes fails to detect some open-world objects that a human may find to be of interest. For example, the grinder on the left, the window at the top-right, etc., are not detected by the network. 
The colored captions on the right side of Fig.~\ref{fig:append_masked_caption} show that our approach sometimes fails to generate meaningful object-centric captions for small objects. For the purple object (cushion on a sofa), the caption (`the the the ...') is not meaningful since it fails to form a complete sentence or capture the identity of the object it represents. For the red object (other cushions on a sofa), the caption (`a family sits on the couch') is not object-centric since it fails to provide a description specific to the object. Fig.~\ref{fig:limitation} further highlights a failure mode. After a train crosses the scene for a prolonged period ($\sim30$ frames),  object identities may be lost.

These issues can be addressed by stronger strategies for open-world object discovery, stronger caption-generators, and by integrating better object trackers, which we leave for future work.


\end{document}